\documentclass[letterpaper, 10 pt, conference]{ieeeconf}
\usepackage{cite} 
\usepackage{multirow}
\usepackage[utf8]{inputenc}
\usepackage[misc]{ifsym}
\usepackage{graphicx}
\usepackage{tikz}
\usepackage{algorithm}
\usepackage{algorithmic}
\usepackage{pifont}
\newcommand{\cmark}{\ding{51}}%
\newcommand{\xmark}{\ding{53}}%
\usepackage[misc]{ifsym}
\usepackage{tikz}
\usepackage{pdfpages}
\usepackage{graphicx}
\usepackage{amsmath}
\usepackage[utf8]{inputenc}
\usetikzlibrary{arrows,shapes,positioning,shadows,trees,calc,decorations.markings}

\usepackage{amssymb}
\usepackage{hyperref}

\makeatletter
\newcommand{\printfnsymbol}[1]{%
  \textsuperscript{\@fnsymbol{#1}}%
}
\makeatother

\IEEEoverridecommandlockouts                

\title{\LARGE \bf
ST-MTL: Spatio-Temporal Multitask Learning Model to Predict Scanpath While Tracking Instruments in Robotic Surgery
}

\author{Mobarakol Islam, Vibashan VS, Chwee Ming Lim, and Hongliang Ren
\thanks{M. Islam is with NUS Graduate School for Integrative Sciences and Engineering (NGS), National University of Singapore, Singapore}
\thanks{M. Islam, V. VS and H. Ren are with Dept. of Biomedical Engineering, National University of Singapore, Singapore; Corresponding author: Hongliang Ren, hlren@ieee.org http://labren.org}
\thanks{V. VS  is with Dept. of Instrumentation and Control Engineering, NIT Trichy, India}
\thanks{C. M. Lim  is with Dept. of Otolaryngology, Singapore General Hospital, Singapore}
}

\begin{document}

\maketitle
\thispagestyle{empty}
\pagestyle{empty}

\begin{abstract}
Representation learning of the task-oriented attention while tracking instrument holds vast potential in image-guided robotic surgery. Incorporating cognitive ability to automate the camera control enables the surgeon to concentrate more on dealing with surgical instruments. The objective is to reduce the operation time and facilitate the surgery for both surgeons and patients. We propose an end-to-end trainable Spatio-Temporal Multi-Task Learning (ST-MTL) model with a shared encoder and spatio-temporal decoders for the real-time surgical instrument segmentation and task-oriented saliency detection. In the MTL model of shared-parameters, optimizing multiple loss functions into a convergence point is still an open challenge. We tackle the problem with a novel asynchronous spatio-temporal optimization (ASTO) technique by calculating independent gradients for each decoder. We also design a competitive squeeze and excitation unit by casting a skip connection that retains weak features, excites strong features, and performs dynamic spatial and channel-wise feature recalibration. To capture better long term spatio-temporal dependencies, we enhance the long-short term memory (LSTM) module by concatenating high-level encoder features of consecutive frames. We also introduce Sinkhorn regularized loss to enhance task-oriented saliency detection by preserving computational efficiency. We generate the task-aware saliency maps and scanpath of the instruments on the dataset of the MICCAI 2017 robotic instrument segmentation challenge. Compared to the state-of-the-art segmentation and saliency methods, our model outperforms most of the evaluation metrics and produces an outstanding performance in the challenge.

\end{abstract}


\section{Introduction}
Image-guided robotic surgery is revolutionizing advancement in minimally-invasive surgery, which increases the precision, flexibility, and repeatability of surgical procedures. For example, Da Vinci telerobotic surgical system \cite{ngu2017vinci} is assisting surgeons to perform complex dexterous manipulations using the master-slave console and stereoscopic vision system. However, the surgeon still needs to perform the manual movement of the endoscope camera, which includes additional workload during the intervention with surgical instruments. It limits the efficiency in robotic surgery by interrupting the surgeon's concentration on the crucial surgical task. Automating camera guidance with the surgeon's visual perception may significantly lessen the operating duration.
Further, there remain challenges in image cognition because of the reduction of instrument size and actuation mechanisms, complicated surgical environment with shadows, adverse lighting, partial occlusion, and smoke \cite{allan20192017}. Therefore, enabling visual perception based endoscope automation while tracking and segmentation of the instruments in robotic surgery is crucial to enhance surgical outcomes. 
 
Visual saliency detection focuses on determining the image regions that attract human visual attention in computational neuroscience \cite{treue2003visual}. The saliency map or attention map represents eye fixation points generated by human observers when exploring an image naturally or task-aware in mind. These fixation points are traditionally captured with eye tracker \cite{judd2009learning} or mouse click \cite{pan2017salgan}. Human visual attention enables surgeons to quickly identify and locate potential risks or important visual cues across the surgical view. This task-oriented saliency map contains relative importance of the different parts of an image, and these are aggregated over the temporal dimension. The scanpath can also represent Task-oriented temporally-aware attention or the surgeon's gaze. It is a temporally-aware saliency representation that has received recent priority in the neuroscience community \cite{assens2018pathgan}. Robot-assisted interventions are highly task-oriented, and it is crucial to understand task-oriented attention to assist surgeons by guiding camera movements during surgery. Besides, the attention map can be used to train the new surgeon \cite{chetwood2012collaborative}, assess the workflow, and guide the surgical camera where to focus, how long, and which order.
 
In recent years, deep neural network is showing success in the tasks of segmentation \cite{islam2018glioma}, detection \cite{redmon2016you}, tracking \cite{li2019siamrpn++}, and visual perception \cite{wang2019revisiting} from computer vision to medical imaging. However, most of these models are designed and optimized for the specific task. This bounds the impact of the deep learning model where the human brain can learn and function multiple tasks at a time. Visual data usually contains shared information on multiple tasks and can be leveraged to learn a common feature representation for all the tasks. For example, surgical instrument tracking and scanpath prediction underlie under the same surgical scene and govern by the same properties and dynamics. By considering these factors, we propose a spatio-temporal multi-task learning (ST-MTL) model for surgical instrument segmentation and saliency detection in this work. Task-aware temporal attention map and corresponding scanpath are generated on the MICCAI 2017 surgical instrument segmentation dataset \cite{allan20192017} to train our ST-MTL model.

\begin{figure*}[!t]
    \centering
    \includegraphics[width=.8\textwidth]{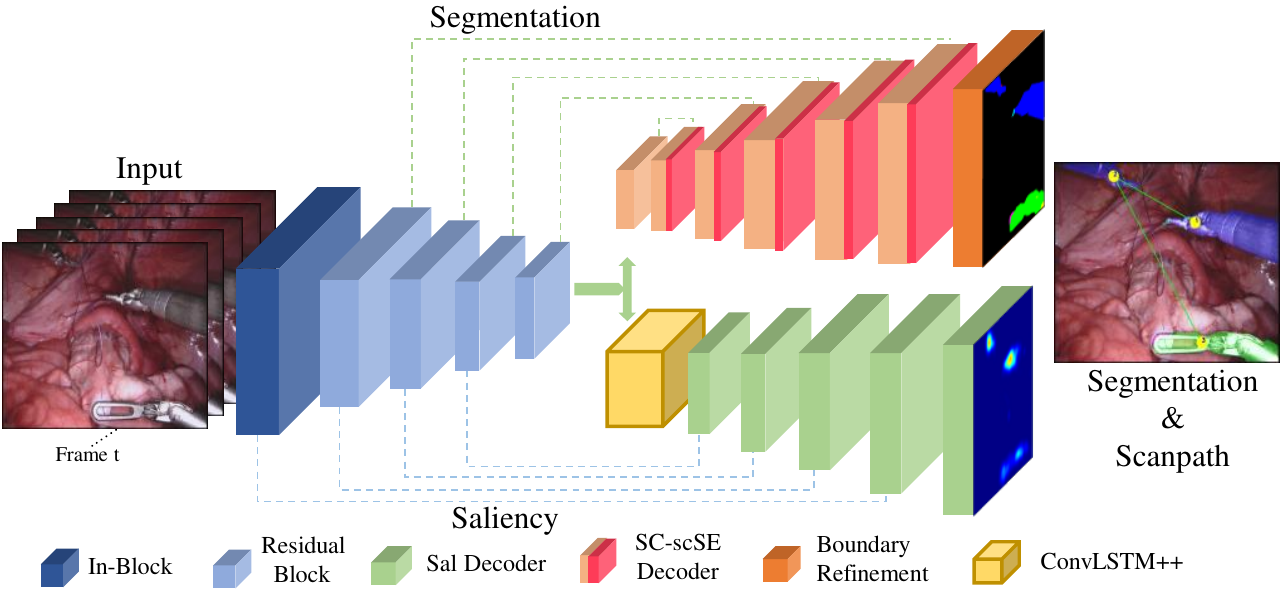}
    \caption{Our proposed ST-MTL model. It has shared encoder and spatio-temporal task-aware decoders for the segmentation and saliency prediction.}
    \label{fig:mtl_model}
\end{figure*}

\subsection{Related Work}
Earlier works in image-guided robotic surgery focus on tool detection, segmentation, and camera automation by considering a single-task model. There are also few gaze-contingent approaches to train novice surgeons or control camera movement with an additional eye-tracking device.  

\subsubsection{Surgical Instrument Tracking} 
Surgical instrument tracking studies using the convolutional neural network (CNN) can be divided into tracking-by-detection and tracking-by-segmentation. For example, tracking-by-detection methods using multi-image fusion CNN \cite{al2017surgical}, region proposal based multimodal network \cite{sarikaya2017detection}, pre-trained Yolo \cite{choi2017surgical}, and region-based CNN \cite{jin2018tool}. However, detection of the high-resolution surgical video is computationally expensive, and the bounding box is unable to delineate the exact contour of the instrument. On the other hand, there are also several works on instrument tracking-by-segmentation. Auxiliary supervised deep adversarial learning model is developed to segment the instrument in robotic surgery \cite{islam2019real}. LinkNet \cite{chaurasia2017linknet} with  Jaccard index based loss function is used for binary, part, and type-wise instrument segmentation \cite{shvets2018automatic}. Subsequently, a holistically nested CNN approach ToolNet \cite{garcia2017toolnet}, FCN with affine transformation, joint CNN, and recurrent neural network (RNN) are utilized to track the surgical instrument.
Nonetheless, the performance of these models is not satisfactory, especially for the instrument type segmentation \cite{allan20192017}. Moreover, with less computational resources, achieving multiple tasks from a single model is more competent. Concurrent segmentation and pose estimation model is developed to track surgical instrument by extending encoder-decoder segmentation architecture with fully-connected layers for pose point detection \cite{laina2017concurrent}. The approach tries to predict multiple objectives without considering the real-time speed.

\subsubsection{Surgical Camera Automation} 
Recently, the research in surgical camera guidance is one key focus in the field of robotic surgery. There are several strategies to control the endoscope using color marker \cite{probst2017automatic}, head movement \cite{stolzenburg2011comparison}, voice command \cite{munoz2005pivoting}, and learning from demonstration \cite{rivas2019transferring}. The main limitation of these approaches is that they either consider very rigid rules such as the relative distance of the instrument and endoscope or calculate the middle point of the instruments. The preference for endoscope guidance depends on gaze-contingent cognitive knowledge. A gaze-contingent strategy is developed to automate the endoscope movements by tracking the eye gaze of the surgeon \cite{noonan2010gaze}. However, the approach requires an additional eye-tracking device, which may create a distraction in the surgical environment. In our previous work \cite{islam2019learning}, an MTL model is designed to guide camera focus while segmenting surgical instruments. The model obtains significant improvement of segmentation by using a novel design of decoder with spatial and channel \textit{squeeze \& excitation} (scSE) \cite{roy2018concurrent} unit. The main limitation of the approach is that the temporal information is not considered where task-oriented camera guidance mostly depends on long-range activity between consecutive video frames. 

\subsubsection{Saliency detection and Scanpath}
Visual saliency has received interest in the neuroscience research community due to the rise in perceiving human visual perception. In prior works, most of the saliency detection models are inspired by primate visual systems. \textit{Itti \& Koch} \cite{itti1998model} propose a conceptually simple computational model which combine low-level features at multi-scale for saliency-driven focal visual attention. The graph-based saliency model is introduced from low-level feature maps that work upon Markov chains over various image maps \cite{harel2007graph}. In \cite{judd2009learning}, a bottom-up and top-down model of saliency detection is presented by using low, mid, and high-level image features. Recently, deep CNN is applying successfully for attention map detection. DSLSTM model \cite{liu2018deep} forms with CNN and long-short term memory (LSTM) and incorporates global contexts to mimic the human visual system. SalGAN \cite{pan2017salgan} uses adversarial training over a deep CNN with a simple encoder-decoder architecture and binary cross-entropy (BCE) loss as an adversarial loss to estimate the saliency. However, all these models are based on natural saliency, where most of the real-world applications are task-oriented. Task-dependent attention map is generated by cues from both bottom-up visual saliency and high-level attention transition \cite{huang2018predicting}. These output features from CNN are sent to LSTM to predict human attention from previous fixations by exploiting the temporal context of gaze fixations. PathGAN \cite{assens2018pathgan} consists of an end-to-end model capable of predicting scanpaths on ordinary and omnidirectional images using the framework of conditional adversarial networks \cite{isola2017image}. However, the mechanism of fixation sequence or scanpath depends on temporal information, and predicting scanpath from saliency map needs to consider spatio-temporal information with long-range sequences of video frames.

\subsubsection{Multi-task learning (MTL)}
The idea of multi-task learning (MTL) is investigated first for shallow approaches in \cite{caruana1997multitask}, \cite{argyriou2007multi}, \cite{evgeniou2004regularized}. These studies observe that learning multiple tasks with a model can be better than learning them independently. MaskRCNN \cite{abdulla2017mask} is one of the state-of-the-art MTL architecture forms of region proposal network for object detection and FCN decoder, for instance, segmentation. MTL models of joint semantic segmentation and object detection \cite{sistu2019real, dvornik2017blitznet}, segmentation, and depth estimation \cite{nekrasov2018real} are developed using the shared encoder and task-oriented decoders. More recently, UberNet \cite{kokkinos2017ubernet} integrates multiple vision tasks such as semantic segmentation, object detection, boundary estimation into a single deep neural network pipeline. However, these models either heavy-weighted and poorly optimized for multiple tasks. Therefore, current MTL models are unable to apply for the real-time application and obtained inadequate performance.

\subsection{Contributions}
In this paper, we address the challenges of the spatio-temporal MTL model for robotic instrument segmentation and task-oriented saliency detection with asynchronous spatio-temporal optimization. Our contributions are summarized as follows:
\begin{itemize}
    \item[--] We propose a \textit{spatio-temporal MTL (ST-MTL)} model with a weight-shared encoder and task-aware spatio-temporal decoders.
    \item[--] We introduce a novel way to train the proposed MTL model by using \textit{asynchronous spatio-temporal optimization (ASTO)}. 
    \item[--] Our novel design of decoders, \textit{skip competitive scSE unit}, and \textit{ConvLSTM++} boost up the model performance. 
    \item[--] We generate task-oriented instruments saliency and scanpath similar to the surgeon's visual perception to get the priority focus on surgical instruments. Our model achieves impressive results and surpasses the existing state-of-the-art models in MICCAI robotic instrument segmentation dataset.
\end{itemize}

\section{Proposed Approach}
In this work, we propose a novel Spatio-Temporal Multi-Task Learning (ST-MTL) model for segmentation and task-oriented attention, as illustrated in Fig. \ref{fig:mtl_model}. The proposed MTL model forms with a shared encoder and two task-aware spatio-temporal decoders. For the optimization of our ST-MTL model, we propose an asynchronous spatio-temporal optimization (ASTO) technique. We introduce a novel channel and spatial excitation module in the segmentation decoder to encapsulate the global and local receptive fields. To capture temporal correlations, we adopt ConvLSTM and enhance it with additional prior features. Our model is an end-to-end multi-task model suitable to supervise real-time robot-assisted surgery.

\subsection{Asynchronous Spatio-Temporal  Optimization (ASTO)}
\label{asto}
 The challenging part of the MTL model is optimizing multiple tasks in a convergence point (or same epoch) with fusion loss function. This optimization can be related to optimizing a single task with added noise, where the noise correlates the loss from other tasks. Moreover, there is an additional challenge in optimizing spatial and temporal branches of the model simultaneously. The optimization of the temporal branch depends on sequential data, whereas the spatial branch requires to shuffle the input. To overcome these problems, we introduce \textit{Asynchronous Spatio-Temporal Optimization} (ASTO). In this technique, we optimize the task-aware spatial and temporal decoders by calculating the gradient independently. Further, we regularize the ST-MTL model for few epochs with a low learning rate.

\begin{algorithm}[!h]
\caption{\small{Asynchronous Spatio-Temporal Optimization}}
\begin{algorithmic}[1]
\label{algortihm_asto}
\small
\STATE \textbf{Initialize model weights}:\\shared (${W_{sh}}$), spatial (${W_s}$),  temporal (${W_t}$)\\
\STATE \textbf{Set gradient accumulators to zero:}\\shared (${dW_{sh}}$), spatial (${dW_s}$),  temporal (${dW_t}$)\\
 $\mathbf{dW_{sh}} \leftarrow 0,\; \mathbf{dW_{s}} \leftarrow 0,\; \mathbf{dW_{t}} \leftarrow 0$\\
\STATE \textit{[Spatial optimization]}\\
 $\mathbf{while}\; spatial\; task\; not\; converged\; \mathbf{do}:$ \\
\hspace{0.25cm}\textit{[Shared encoder gradients w.r.t spatial loss ${\L_{s}}$]}\\
 \hspace{0.25cm}$\mathbf{dW_{sh}} \leftarrow \mathbf{dW_{sh}} + \sum_{i}\delta_{i}\nabla_{W_{sh}}\L_{s}(W_{sh},W_{s})$\\
 \hspace{0.25cm}\textit{[Spatial Task gradients w.r.t spatial loss ${\L_{s}}$]}\\
 \hspace{0.25cm}$\mathbf{dW_{s}} \leftarrow \mathbf{dW_{s}} + \sum_{i}\delta_{i}\nabla_{W_{s}}\L_{s}(W_{sh},W_{s})$\\
 $\mathbf{end \; while}$\\
\STATE \textit{[Temporal optimization]}\\
 $\mathbf{while}\; temporal\; task\; not\; converged\; \mathbf{do}:$ \\
 \hspace{0.25cm}\textit{[Temporal Task gradients w.r.t temporal loss ${\L_{t}}$]}\\
\hspace{0.25cm}$\mathbf{dW_{t}} \leftarrow \mathbf{dW_{t}} + \sum_{i}\delta_{i}\nabla_{W_{t}}\L_{t}(W_{sh}, W_{t})$\\
 $\mathbf{end \; while}$\\

\STATE \textit{[Spatio-Temporal regularization]}\\
 $\mathbf{while}\; both\; tasks\; improving\; \mathbf{do}:$ \\
\hspace{0.25cm}$\mathbf{dW_{sh}} \leftarrow \mathbf{dW_{sh}} + \sum_{i}\delta_{i}\nabla_{W_{sh}}\L(W_{sh},W_{s},W_{t})$\\
\hspace{0.25cm}$\mathbf{dW_{s}} \leftarrow \mathbf{dW_{s}} + \sum_{i}\delta_{i}\nabla_{W_{s}}\L(W_{sh},W_{s},W_{t})$\\
\hspace{0.25cm}$\mathbf{dW_{t}} \leftarrow \mathbf{dW_{t}} + \sum_{i}\delta_{i}\nabla_{W_{t}}\L(W_{sh},W_{s},W_{t})$\\
 $\mathbf{end \; while}$\\
\end{algorithmic}
\end{algorithm}

Algorithm \ref{algortihm_asto} describes our ASTO technique. We do Xavier initialization and construct mini-batches for training. The data-loader is designed to load shuffle and sequence data during spatial and temporal optimization, respectively. First, a shared encoder with a spatial task-aware (segmentation) decoder is optimized by freezing temporal decoder (saliency). Then temporal decoder is optimized with sequential data input by freezing shared encoder and spatial decoder. The gradients are calculated with the corresponding task loss in each phase. Finally, to generalize the encoder's shared weights, we opt end-to-end regularization by optimizing both trained decoders with a small learning rate of \(10^{-5}\). This approach ensures the MTL model to be more generalized by regularizing the overall gradient flow of the network. In our ST-MTL, spatial and temporal decoders are segmentation and task-oriented saliency, respectively.

\subsection{Skip Competitive scSE (SC-scSE)}

\begin{figure}[!h]
    \centering
    \includegraphics[width=0.4\textwidth]{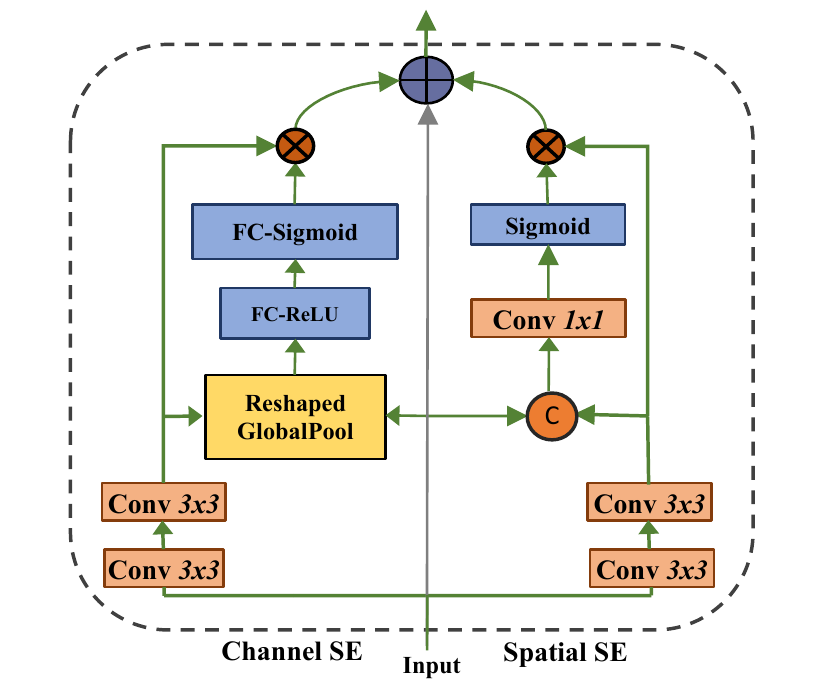}
    \caption{The architecture of the skip competitive scSE (SC-scSE) module. We incorporate inner-imaging in both channel and spatial excitation to increase model performance. After excitation, casting a skip connection to retain the weak feature, which reduces the sparsity of feature learning in deeper layers.}
    \label{fig:SCscSE}
\end{figure}

Spatial and channel "Squeeze \& Excitation" (scSE) \cite{roy2018concurrent, hu2018squeeze} and competitive inner-imaging scSE \cite{hu2018competitive} recalibrates the feature maps to suppress the weak features and signify the essential features. However, squeezing irrelevant features towards zero increases the sparsity in deeper layers and limits the parameter learning \cite{uhrig2017sparsity}. In \cite{orhan2017skip}, skip connection is used to eliminate the singularity and sparsity in CNN learning. To resolve this issue, we also add skip connection, which retains the weak features and enhances the useful features by boosting excitation. We propose to skip competitive scSE (SC-scSE) by adding skip connection and competitive inner-imaging unit into scSE, as in Fig. \ref{fig:SCscSE}. In SC-scSE, the output of channel excitation $\tilde{\mathrm{x}}_{cE}$ and spatial excitation $\tilde{\mathrm{x}}_{sE}$ are integrated with skip input $\mathrm{x}_{r}$. Therefore, the output excited feature maps $\mathrm{x}_{SC}$ can be formulated as given in equation \ref{equ:SC-scSE},

\begin{figure}[!h]
    \centering
    \includegraphics[width=0.45\textwidth]{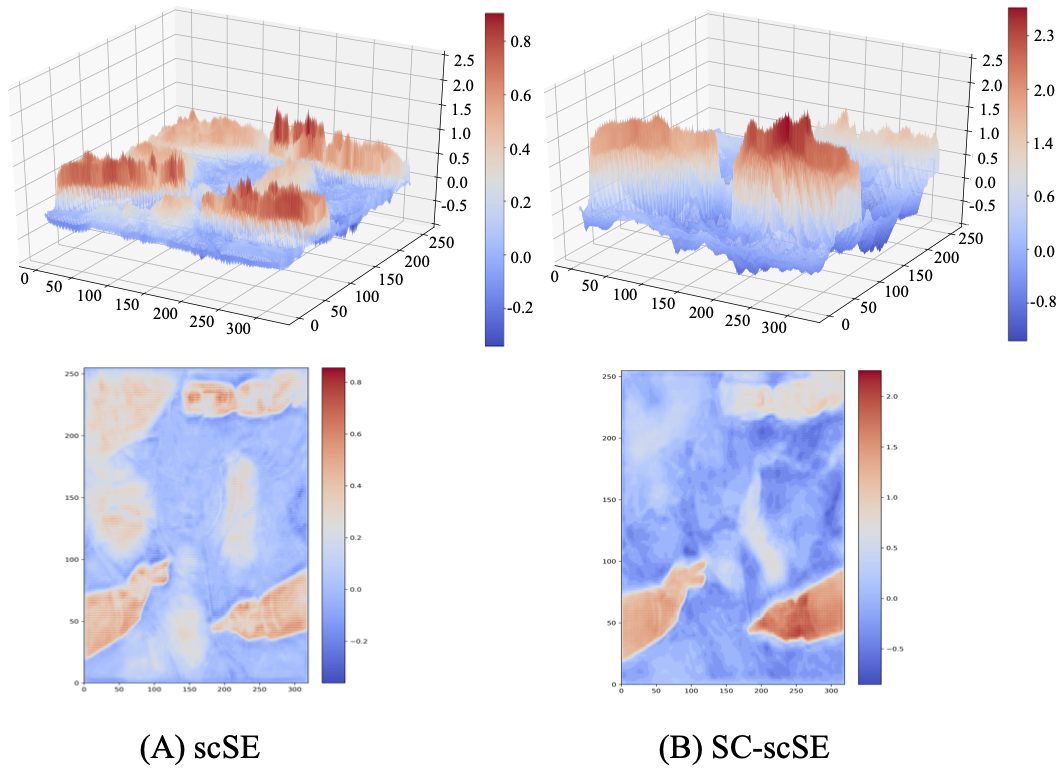}
    \caption{The comparison of the similar output score maps (upper row) and corresponding feature maps (down row) of SC-scSE and scSE. In scSE, output score map values range between \([-0.2,1.0]\) and the range change to \([-1,2.5]\) with SC-scSE. Qualitatively, we can infer that SC-scSE shows higher influence to suppress weak features and excite the strong features comparing to scSE. This characteristic helps to get a well differentiable multi-class confidence map in output prediction.}
    \label{fig:qual_scse}
\end{figure}

\begin{eqnarray}
\label{equ:SC-scSE}
\mathrm{x}_{SC} &=& \tilde{\mathrm{x}}_{cE} + \tilde{\mathrm{x}}_{sE} + \mathrm{x}_{r} 
\end{eqnarray}

where \(\tilde{\mathrm{x}}_{cE}, \tilde{\mathrm{x}}_{sE}, \mathrm{x}_{r}\) corresponds to channel excitation, spatial excitation and skip connection respectively.

In Fig. \ref{fig:qual_scse}, we compare scSE and SC-scSE for similar output feature map. In scSE, when score maps get excited through sigmoid, it tries to differentiate weak feature with the significant feature by exciting the useful feature and suppressing weak feature towards 0. This leads to sparsity and singularity of feature learning in deeper layers of the network. Therefore, incorporating a skip connection resolves these problems by retaining the weak feature and exciting the useful feature. SC-scSE can differentiate weak and useful features dynamically, which eliminates sparsity and singularity in the network. From Fig. \ref{fig:qual_scse}, we can observe that output excitation map of scSE range of \([-0.2,1.0]\) whereas SC-scSE increase the range to \([-1,2.5]\). This characteristic helps to get a well differentiable multi-class confidence map in output prediction. Thus, SC-scSE improves the representational power of a network and performs dynamic spatial and channel-wise feature recalibration by remarkably discriminating weak and significant features. This is why the feature maps with SC-scSE show better calibration over scSE for instruments with background pixels (second row in Fig \ref{fig:qual_scse}). 

\subsection{ConvLSTM++}
\begin{figure}[!h]
    \centering
    \includegraphics[width=0.45\textwidth]{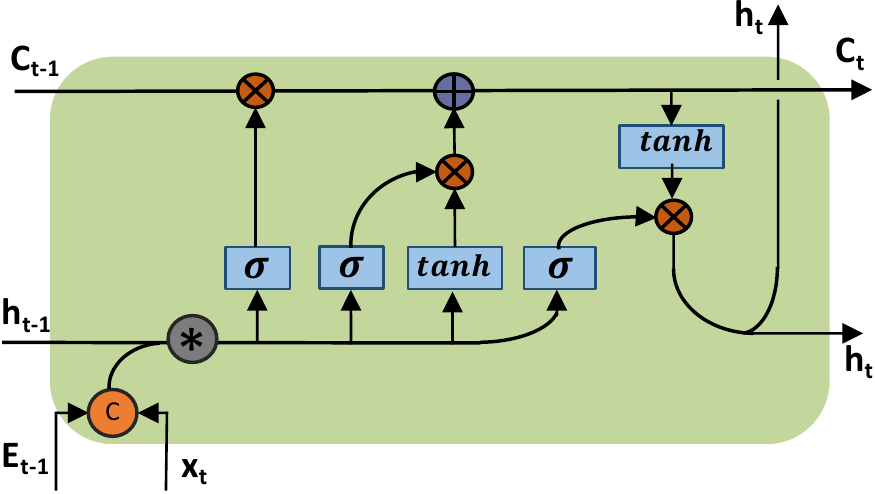}
    \caption{ConvLSTM++ unit network arhitecture. Input gate, output gate, forget gate are crucial for temporal information passage. To learn better spatio-temporal correlation we concatenate high level encoder feature of present frame (\(X_{in}\)) with  previous frame (\(X_{e_{t-1}}\)).}
    \label{fig:lstm}
\end{figure}

In vision-assisted robotic surgery, it is difficult to estimate the task-oriented saliency map and scanpath for the instrument with static frames. Thus, a task-oriented saliency map for the instrument can be improved by ConvLSTM \cite{xingjian2015convolutional}, as it can capture spatio-temporal motion patterns of the instruments. We introduce ConvLSTM++ to understand better long-range dependencies by concatenating high level encoder feature of present frame (\(X_{in}\)) with  previous frame (\(X_{e_{t-1}}\)) as shown in Fig. \ref{fig:lstm}. Our ConvLSTM++ can be formulated as follows,

\begin{eqnarray}
X_{t}&=& X_{in} \oplus X_{e_{t-1}}
\end{eqnarray}

\begin{eqnarray*}
i_{t}&=& \sigma(W_{xi}*X_{t} +  W_{hi}*H_{t-1} + b_{i})
\\ f_{t}&=& \sigma(W_{xf}*X_{t} +  W_{hf}*H_{t-1} + b_{f})
\\ C_{t}&=& f_{t} \circ C_{t-1} + i_{t} \circ \tanh(W_{xc}*X_{t} +  W_{hc}*H_{t-1} + b_{c})
\\ o_{t}&=& \sigma(W_{xo}*X_{t} +  W_{ho}*H_{t-1} + b_{o})
\\ H_{t}&=& o_{t} \circ \tanh(C_{t})
\end{eqnarray*}

\begin{figure}[!h]
\centering
  \includegraphics[width=0.9\linewidth]{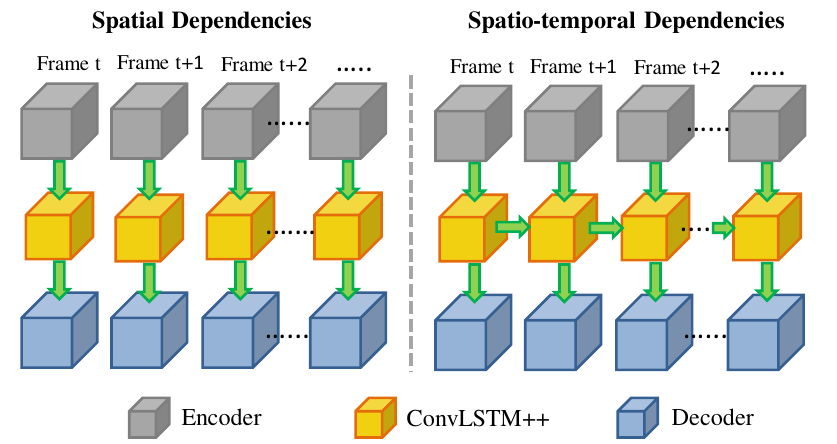}
  \caption{Information cascading between spatial  (left) and spatio-temporal (right) networks.}
  \label{fig:st_flow}
\end{figure}

In equation \((2)\), concatenation increase parameter learning subsequently escalate task-oriented saliency detection. The input gate(\(i_{t}\)), output gate(\(o_{t}\)) and forget gate(\(f_{t}\)) controls the flow of spatio-temporal information flow. The Hadamard product (\( \circ\)) between output gate(\(o_{t}\)) and memory gate(\(C_{t}\)) result in hidden state (\(H_{t}\)) which is crucial for learning long term dependencies. Spatio-temporal feature correlation learning as function of previous time frames is shown in Fig. \ref{fig:st_flow}.

\subsection{Loss Function}
We design  a computational efficient loss function for the task-oriented saliency optimization using binary-cross entropy loss (\textit{BCE}) and Sinkhorn distance (\textit{Sd}) \cite{cuturi2013sinkhorn}, \cite{luise2018differential}. Sd is the entropy regularized version of the Wasserstein distance (\textit{Wd}). 
\textit{BCE} calculates cross-entropy between binary probability distributions, whereas \textit{Sd} calculates the dissimilarity of the probability distribution. The proposed saliency of loss of combined \textit{BCE} ($L_\textrm{bce}$) and \textit{Sd} ($L_\textrm{Sd}$) with weight factor of $\alpha$ ($\alpha = 0.3$ after tuning) can be formulated as-

\begin{equation}
    \mathcal{L}_{\textrm{Ssal}} = \alpha* L_\textrm{Sd} + (1-\alpha)* L_\textrm{bce}
\end{equation}

To obtain optimal computation, we down-sample the ground-truth and prediction maps to \textit{1/4} and measure the distance of a single point to a family of points using matrix-matrix products. If entropy regularization is $\varepsilon*H(x)$ (where $\varepsilon > 0$), to measure the dissimilarity of two distributions of $x$ and $y$, we can write the Sd as \cite{cuturi2013sinkhorn}-

\begin{equation}
    Sd(\mu, \nu) = \inf_{\gamma \in \Pi(\mu, \nu)} {}_{(x, y) \sim \gamma} \big[\:\|x - y\|\:\big] - \varepsilon*H(x)
\end{equation}
On the other hand, multi-class cross-entropy loss is calculated to optimize the segmentation task. In the end-to-end regularization phase (See the detail in section \ref{asto}), a simple fusion of saliency and segmentation loss is used with a low learning rate of \(10^{-5}\).

\subsection{Network Architecture}

\begin{figure}[!h]
    \centering
    \includegraphics[width=0.45\textwidth]{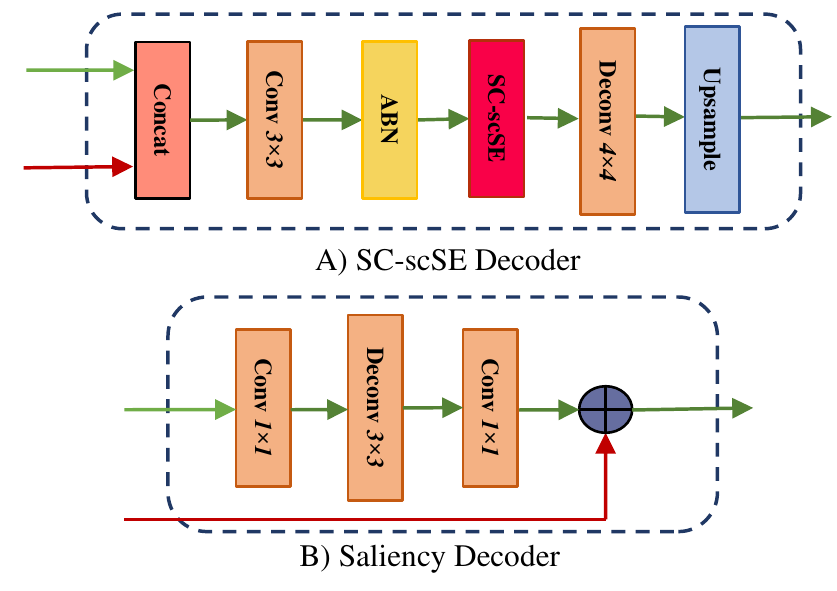}
    \caption{ SC-scSE Decoder for segmentation and Saliency Decoder for task-oriented saliency detection. Placing SC-scSE unit before deconvolution layer facilitates feature learning and produces best segmentation.}
    \label{fig:decoder}
\end{figure}

Our proposed MTL model consists of a weight-shared encoder and task-aware decoders for segmentation and saliency detection, as shown in Fig. \ref{fig:mtl_model}. We adopt a light-weight encoder, ResNet18 \cite{he2016deep} and develop SC-scSE decoder and saliency decoder for segmentation and saliency detection as shown in Fig.\ref{fig:decoder}. Each SC-scSE decoder block concatenates the previous layer output and the output feature of the corresponding encoder block. It passes through the sequential layers of convolution, ABN \cite{li2018adaptive}, SC-scSE, and deconvolution. We observe that placing SC-scSE module before the deconvolution layer improves the feature learning. This is because deconvolution or transpose convolution operation increases the sparsity in network learning, limiting the functionality of the SC-scSE module. The prediction maps of the segmentation decoder are refined by exploiting a boundary refinement module as \cite{peng2017large}. In saliency decoder, encoder feature maps of frame \textit{t} and \textit{t-1} are concatenated and fed into ConvLSTM++ to obtain better long-range spatio-temporal correlation. The output of ConvLSTM++ propagates to saliency decoder blocks containing sequential convolution, deconvolution, and convolution layers (illustrated in Fig. \ref{fig:decoder}B). The decoder's generated output is fused with corresponding encoder feature maps. The final saliency map is obtained by applying sigmoid over the output of the saliency decoder.

\begin{figure*}[!h]
    \centering
    \includegraphics[width=1\textwidth]{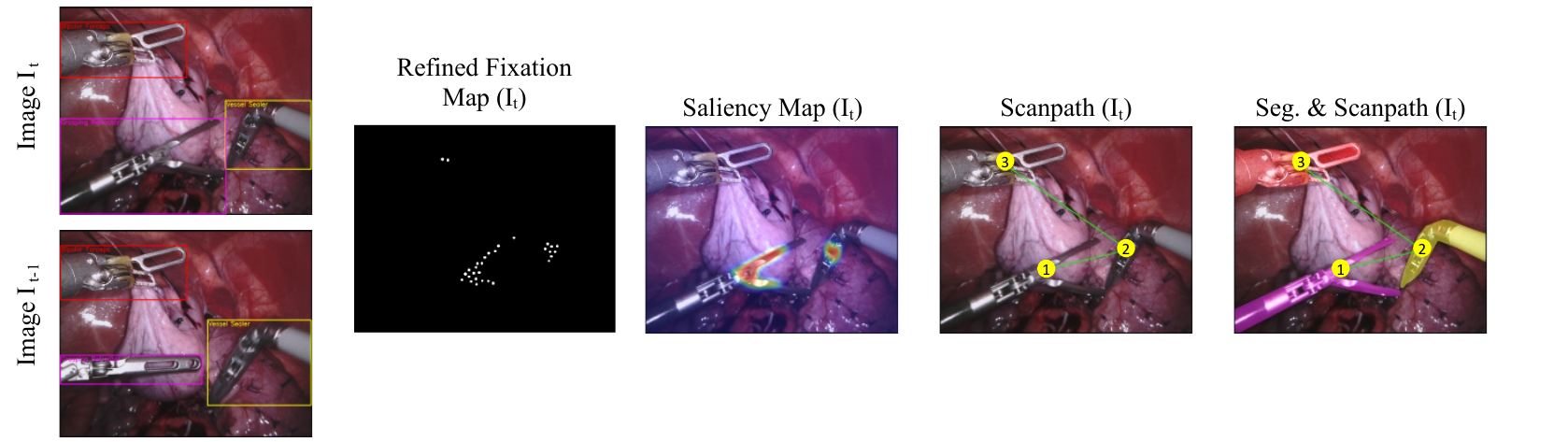}
    \caption{Generation of saliency map and scanpath for surgical instruments using weighted fuction of instruments movement and size. We only consider displacement and deformation of the instruments in consecutive frames to calculate scanpath.}
    \label{fig:scanpath_gen}
\end{figure*}

ResNet18 \cite{he2016deep} as light-weight encoder network and proposed SC-scSE decoder and saliency decoder for segmentation and task-oriented saliency detection as shown in Fig.\ref{fig:decoder}. SC-scSE decoder concatenates the previous layer output and skip features of the corresponding encoder block and passes through the sequential layers of convolution, ABN (Adaptive Batch Normalization), SC-scSE and deconvolution. We observe that placing SC-scSE module before the deconvolution layer improves the feature learning. This is because deconvolution or transpose convolution operation increases the sparsity in network learning, limiting the functionality of the SC-scSE module. The prediction maps of the segmentation decoder are refined by exploiting a boundary refinement module as \cite{peng2017large}. In saliency decoder, encoder feature maps of frame \textit{t} and \textit{t-1} are concatenated and fed into ConvLSTM++ to obtain better long-range spatio-temporal correlation. The output of ConvLSTM++ propagates to saliency decoder blocks containing sequential convolution, deconvolution, and convolution layers (illustrated in Fig. \ref{fig:decoder}B). The decoder's generated output is fused with corresponding encoder feature maps. The final saliency map is obtained by applying sigmoid over the output of the saliency decoder.

\section{Experiments}

 \begin{table*}[!h]
\centering
\caption{Evaluation score for the cross-validation dataset. The evaluation metrics are dice, Hausdorff(Hausd.), binary cross-entropy loss (BCE), area under curve-Borji (AUC-B) \cite{bylinskii2015saliency}, scanpath order accuracy, and frame per second (FPS). FPS is calculated by a single RTX 2080 Ti GPU with a batch size of 1. The best values of each metric are boldened, and the values better than ours are underlined.\newline}
\scalebox{.94}{\begin{tabular}{l|c|c|c|c|c|c|c|c|c|c}
\hline
\multicolumn{1}{c|}{\multirow{1}{1cm}{{Model}}} & \multicolumn{4}{c|}{{Segmentation}} & \multicolumn{4}{c|}{{Saliency}} & \multirow{1}{.75cm}{{FPS}} &  \multirow{1}{1.1cm}{{Params}}\\ \cline{2-9}
\multicolumn{1}{c|}{} & \multicolumn{2}{c|}{{Binary}} & \multicolumn{2}{c|}{{Type}} & \multicolumn{2}{c|}{{Saliency}} & \multicolumn{2}{c|}{{Scanpath}} &  &\\ \cline{2-9}
\multicolumn{1}{c|}{{}} & {Dice} & {Hausd.} & {Dice} & {Hausd.} & {BCE} & {AUC-B} & {Top-1} & {Avg.} & {} &\\ \hline
Ours & \textbf{0.961} & \textbf{9.89} & \textbf{0.773} & 10.67 & \textbf{0.032} & 0.882 & \textbf{0.715} & \textbf{0.693} & 42 & 51.5M\\ \hline
SalGAN \cite{pan2017salgan}  & - & - & - & - & 0.034 & \underline{0.923} & 0.602 & 0.630 & 223 & 26.7M\\ \hline
DSCLRCN \cite{liu2018deep} & - & - & - & - & 0.033 & 0.730 & 0.621 & 0.654 & 107 & 33.5M\\ \hline
\cite{itti1998model}& - & - & - & - & 0.046 & 0.699 & 0.485 & 0.535 & - & -\\ \hline
\cite{islam2019real} & 0.946  & 9.90  & 0.634 & 14.32 & - & - & - & - & 144 & 14.9M\\ \hline
\cite{peng2017large}& 0.940 & 10.48 & 0.612 & 13.98 & - & - & - & - & 83 & 23.9M\\ \hline
LinkNet & 0.941 & 10.39 & 0.578 & 11.88 & - & - & - & - & 111 & 11.5M\\ \hline
ERFNet\cite{romera2017erfnet} & 0.923 & 11.05 & 0.495 & 11.55 & - & - & - & - & 44 & 2.02M\\ \hline
DeepLabV3+ & 0.947 & 10.07 & 0.657 & \underline{10.47} & - & - & - & - & 56 & 59.3M\\ \hline
scSE U-Net\cite{roy2018concurrent} & 0.931 & 10.09 & 0.542 & 11.88 & - & - & - & - & 107 & 42.7M\\ \hline
TernausNet11 & 0.920 & 11.25 & 0.446 & 12.29 & - & - & - & - & \underline{295} & 34.5M\\ \hline
BiseNet\cite{yu2018bisenet} & 0.942 & 9.99 & 0.440 & 10.80 & - & - & - & - & 54 & 90.8M\\ \hline
\end{tabular}}
\label{table:ours}
\end{table*}

\subsection{Dataset}
This work is done using the robotic instrument segmentation dataset of MICCAI endoscopic vision challenge \cite{allan20192017}. This dataset contains 10 sequences recorded with a resolution of 1920 x 1080 using da Vinci surgical systems \cite{ngu2017vinci}. In each sequence, significant instrument motion and visibility are observed and sampled at a rate of 1 Hz. There are 7 different robotic surgical instruments, such as Large Needle Driver, Cadiere Forceps, Bipolar Forceps, Vessel Sealer, Prograsp Forceps, Monopolar Curved Scissors, and additionally a drop-in ultrasound probe.
For our work, the training data is cropped to 1280 x 1024 by removing the black canvas and split into training, validation. The training set consists of sequences 1, 2, 3, 5, 6, 8, and cross-validation data consist of sequences 4, 7, and for the testing, we follow the challenge regulation to evaluate with all the sequences. To get the evaluation score, we submit our model prediction in the challenge website \footnote[1]{https://endovissub2017-roboticinstrumentsegmentation.grand-challenge.org/}. To produce a ground-truth saliency map, we collect eye gazes as fixation points from an experienced da Vinci robotic surgeon using an eye tracker and convert the points into heatmap using \textit{Multivariate Normal Distribution} \cite{li2009visual, klein2012salient}. Further, we re-weight the fixation points by considering surgical task-oriented factors such as displacement and deformation of the instruments between consecutive frames \cite{islam2019learning}. The scanpath draws based on the density of the fixation points or heatmap for each instrument. An example of scanpath generation from consecutive frames is demonstrated in Fig. \ref{fig:scanpath_gen}.

\begin{figure*}[!h]
    \centering
    \includegraphics[width=1\textwidth]{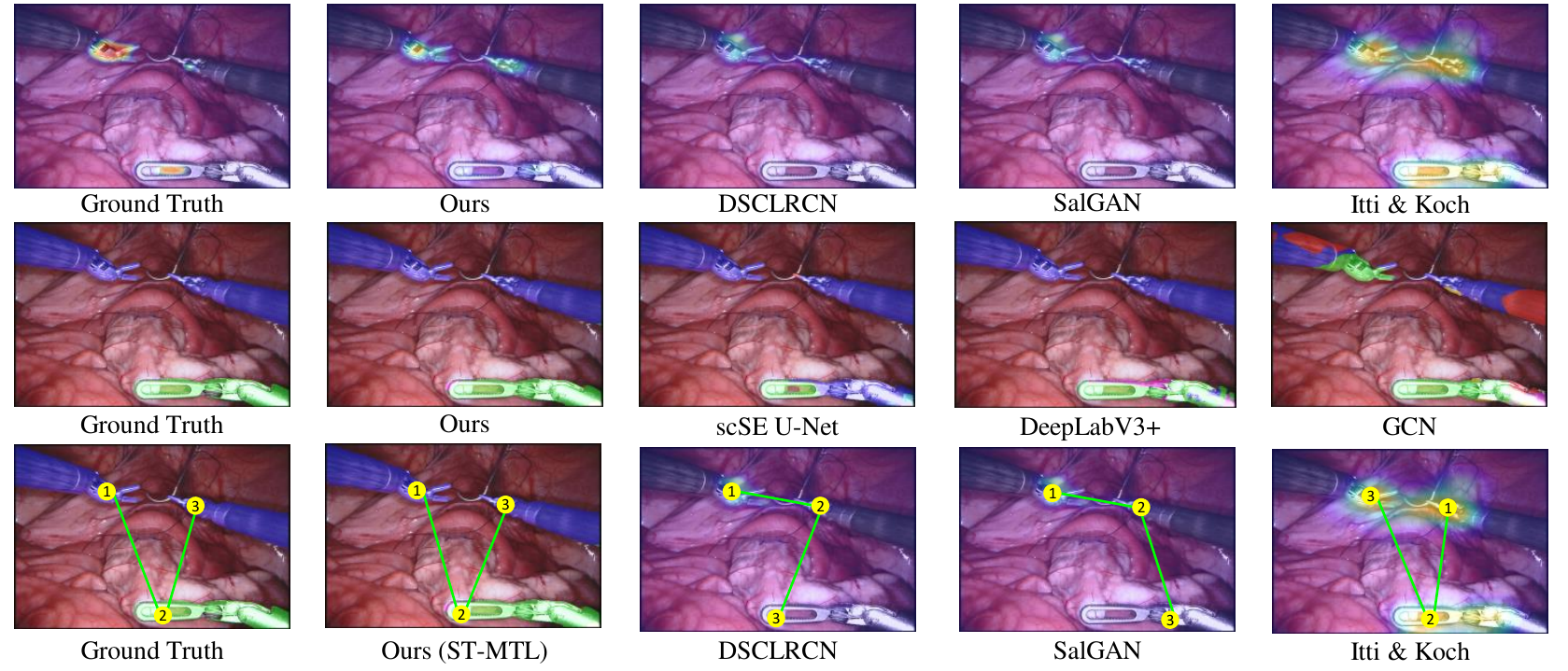}
    \caption{ Visualization of the predicted saliency, segmentation, and scanpath. Our model's prediction is close to the ground-truth where other SOTA models produce a false positive or false negative with the wrong scanpath order. We observe that task-Oriented saliency is crucial for surgical scanpath prediction when compared with natural saliency models like Itti \& Koch.}
    \label{fig:our_comp}
\end{figure*}

\subsection{Implementation Details}
Standard data augmentation is adopted for the training. All images are randomly flipped and normalized by subtracting the mean and divide by standard deviation per channel. For sequence data loading, we set clip length as \(14\) frames and shuffle randomly in a sequence. For other hyper-parameters, we use Adam optimizer with an initial learning rate of 0.0001 and a `poly' learning rate with the power of 0.9 to update it. The momentum and weight decay set constant to 0.99 and $10^{-4}$, respectively. The encoder is initialized with the ImageNet pre-trained model of ResNet18. We exploit our proposed ASTO optimization technique and adopt sequence data loading during regularization. The proposed model is implemented using Pytorch. All the experiments are conducted with Nvidia RTX 2080 Ti GPUs.

\section{Results}
To evaluate our ST-MTL model, we exploit several well-known evaluation metrics of segmentation, saliency, and scanpath ranking.  Dice and Hausdorff are used to measure segmentation, and binary cross-entropy loss(BCE), curve-Borji (AUC-B) \cite{bylinskii2015saliency} are applied to calculate saliency prediction. For scanpath, compare the top rank and all average instruments priority between ground-truth and prediction. Overall evaluation is done by using cross-validation and testing dataset of MICCAI endoscopic vision challenge \cite{allan20192017}. 
 
\begin{figure}[!h]
    \centering
    \includegraphics[width=0.5\textwidth]{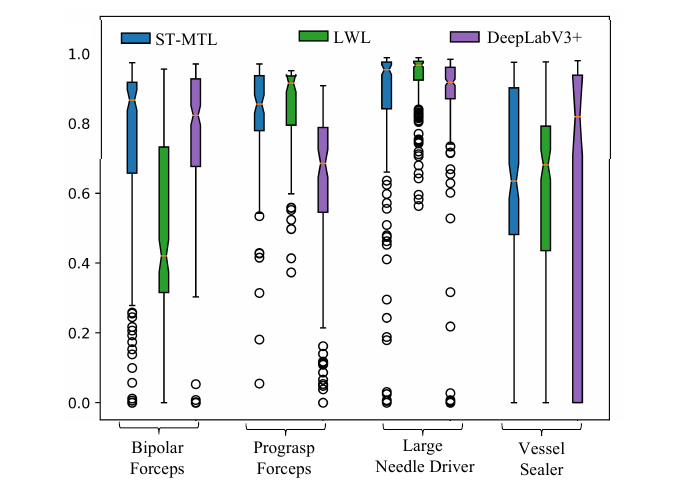}
    \caption{Class-wise boxplot for three best performing models such as Ours, LWL \cite{islam2019real} and DeeplabV3+ \cite{chen2018encoder} in cross-validation. For interpretation of the references to color corresponds to each instrument type prediction post-process coloring.}
    \label{fig:boxplot}
\end{figure}

\begin{table*}[!h]
\centering
\caption{The comparative evaluation of the binary segmentation prediction of between our ST-MTL and participated models in the challenge \cite{allan20192017}. We produce the segmentation by following challenge regulations and evaluated by challenge portal. The best scores of each metric are boldened. The top scores in the challenge are underlined.}
\scalebox{1}{\begin{tabular}{l|c|c|c|c|c|c|c|c|c|c|l}
\hline
\textbf{Model} & \multicolumn{1}{l|}{\textbf{Seq\_1}} & \textbf{Seq\_2} & \multicolumn{1}{l|}{\textbf{Seq\_3}} & \multicolumn{1}{l|}{\textbf{Seq\_4}} & \multicolumn{1}{l|}{\textbf{Seq\_5}} & \multicolumn{1}{l|}{\textbf{Seq\_6}} & \multicolumn{1}{l|}{\textbf{Seq\_7}} & \multicolumn{1}{l|}{\textbf{Seq\_8}} & \multicolumn{1}{l|}{\textbf{Seq\_9}} & \multicolumn{1}{l|}{\textbf{Seq\_10}} & \textbf{Mean} \\ \hline
\textbf{Ours} & 0.805 & \textbf{0.826} & \textbf{0.978} & \textbf{0.968} & \textbf{0.892} & \textbf{0.955} & 0.913 & \textbf{0.953} & \textbf{0.886} & \textbf{0.927} & \textbf{0.910} \\ \hline
\textbf{MIT} & \underline{ 0.854} & 0.794 & \underline{ 0.949} & \underline{ 0.949} & \underline{ 0.862} & \underline{ 0.922} & 0.856 & \underline{ 0.937} & 0.865 & 0.905 & \underline{ 0.888} \\ \hline
\textbf{NCT} & 0.784 & 0.788 & 0.926 & 0.934 & 0.701 & 0.876 & 0.846 & 0.881 & 0.789 & 0.899 & 0.843 \\ \hline
\textbf{UB} & 0.807 & \underline{ 0.806} & 0.914 & 0.925 & 0.740 & 0.890 & \underline{ 0.930} & 0.904 & 0.855 & \underline{ 0.917} & 0.875 \\ \hline
\textbf{SIAT} & 0.625 & 0.669 & 0.897 & 0.907 & 0.604 & 0.843 & 0.832 & 0.513 & 0.839 & 0.899 & 0.803 \\ \hline
\textbf{UCL} & 0.631 & 0.645 & 0.895 & 0.883 & 0.719 & 0.852 & 0.710 & 0.517 & 0.808 & 0.869 & 0.785 \\ \hline
\textbf{UA} & 0.413 & 0.463 & 0.703 & 0.751 & 0.375 & 0.667 & 0.362 & 0.797 & 0.539 & 0.689 & 0.591 \\ \hline
\textbf{BIT} & 0.275 & 0.282 & 0.455 & 0.310 & 0.220 & 0.338 & 0.404 & 0.366 & 0.236 & 0.403 & 0.326 \\ \hline
\textbf{TUM} & 0.760 & 0.799 & 0.916 & 0.915 & 0.810 & 0.873 & 0.844 & 0.895 & \underline{ 0.877} & 0.909 & 0.873 \\ \hline
\textbf{Delhi} & 0.408 & 0.524 & 0.743 & 0.782 & 0.528 & 0.292 & 0.593 & 0.562 & 0.626 & 0.715 & 0.612 \\ \hline
\textbf{UW} & 0.337 & 0.289 & 0.483 & 0.678 & 0.219 & 0.619 & 0.325 & 0.506 & 0.377 & 0.603 & 0.461 \\ \hline
\end{tabular}}
\label{table:chall_bin}
\end{table*}

\begin{table*}[!h]
\centering
\caption{The comparative evaluation of the type segmentation prediction of between our ST-MTL and participated models in the challenge \cite{allan20192017}. We produce the segmentation by following challenge regulations and evaluated by challenge portal. The best scores of each metric are boldened. The top scores in the challenge are underlined.}
\scalebox{1}{\begin{tabular}{l|c|c|c|c|c|c|c|c|c|c|l}
\hline
\textbf{Model} & \multicolumn{1}{l|}{\textbf{Seq\_1}} & \textbf{Seq\_2} & \multicolumn{1}{l|}{\textbf{Seq\_3}} & \multicolumn{1}{l|}{\textbf{Seq\_4}} & \multicolumn{1}{l|}{\textbf{Seq\_5}} & \multicolumn{1}{l|}{\textbf{Seq\_6}} & \multicolumn{1}{l|}{\textbf{Seq\_7}} & \multicolumn{1}{l|}{\textbf{Seq\_8}} & \multicolumn{1}{l|}{\textbf{Seq\_9}} & \multicolumn{1}{l|}{\textbf{Seq\_10}} & \textbf{Mean} \\ \hline
\textbf{Ours} & \textbf{0.276} & \textbf{0.830} & \textbf{0.931} & \textbf{0.951} & 0.492 & 0.501 & \textbf{0.480} & 0.707 & \textbf{0.409} & \textbf{0.832} & \textbf{0.640} \\ \hline
\textbf{MIT} & \underline{0.177} & \underline{0.766} & 0.611 & \underline{0.871} & \underline{0.649} & \underline{0.593} & 0.305 & \underline{0.833} & \underline{0.357} & 0.609  & \underline{0.542} \\ \hline
\textbf{NCT} & 0.056 & 0.499 & \underline{0.926} & 0.551 & 0.442 & 0.109 & 0.393 & 0.441 & 0.247 & 0.552  & 0.409 \\ \hline
\textbf{UB} & 0.111 & 0.722 & 0.864 & 0.680 & 0.443 & 0.371 & \underline{0.416} & 0.384 & 0.106 & 0.709  & 0.453 \\ \hline
\textbf{SIAT} & 0.138 & 0.013 & 0.537 & 0.223 & 0.017 & 0.462 & 0.102 & 0.028 & 0.315 & \underline{0.791}  & 0.371 \\ \hline
\textbf{UCL} & 0.073 & 0.481 & 0.496 & 0.204 & 0.301 & 0.246 & 0.071 & 0.109 & 0.272 & 0.583  & 0.337 \\ \hline
\textbf{UA} & 0.068 & 0.244 & 0.765 & 0.677 & 0.001 & 0.400 & 0.000 & 0.357 & 0.040 & 0.715  & 0.346 \\ \hline
\end{tabular}}
\label{table:chall_type}
\end{table*}

 \subsection{Cross-validation}
 
 Table \ref{table:ours} represents the comprehensive evaluation of our ST-MTL model with the comparison of state-of-the-art (SOTA) models for the tasks of segmentation and saliency. To ensure fair comparison, all the experiments conducted on the same pre-processing and augmentation techniques on dataset. We also use vanilla architecture for most of the methods from authors' recommended or shared Github repository. To initialize the model parameters, we utilize the pre-trained model (with Imagenet, MS coco, or SALICON dataset) if it is available. Our model outperforms other SOTA models with dice accuracy in both binary and instrument type segmentation. Specially, a significant improvement appears in type-wise segmentation with the dice score of \textit{76.8\%} where nearest model is DeepLabV3+ \cite{chen2017rethinking} with \textit{65.7\%}.  However, BiseNet \cite{yu2018bisenet} obtains competitive Hausdorff performance with our model. On the other hand, our model also achieves the best performance in saliency metrics of BCE and scanpath accuracy. Though SalGAN \cite{pan2017salgan} achieves better curve-Borji (AUC-B) score but poor scanpath accuracy. The qualitative results of our model and different SOTA models for the cross-validation dataset is shown in Fig. \ref{fig:our_comp} . We can infer that our model predicts instrument segmentation with less false positive and best saliency and task-oriented scanpath.

 Fig. \ref{fig:boxplot} demonstrates the boxplot of the type-wise prediction for three best performing models such as Ours, LWL \cite{islam2019real} and DeeplabV3+ \cite{chen2017rethinking} in cross-validation. The prediction of some instruments is poor as less existence in the dataset.
 

\subsection{Evaluation with Challenge}

\begin{figure*}[!h]
    \centering
    \includegraphics[width=1\textwidth]{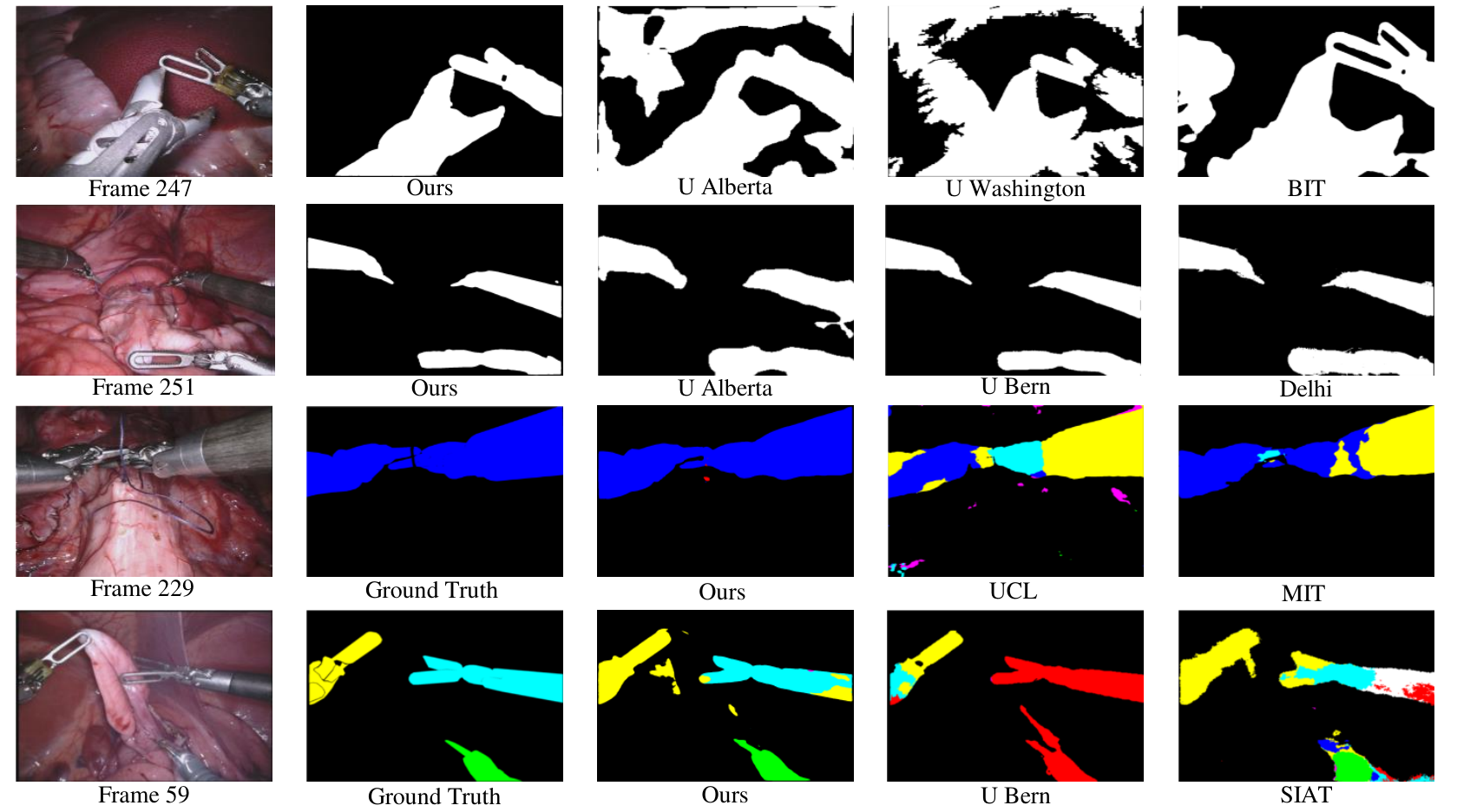}
    \caption{Qualitative results comparison of frames from the different datasets for binary and type segmentation with corresponding results from the methods of robotic instrument segmentation challenge \cite{allan20192017} and proposed ST-MTL.}
    \label{fig:chal}
\end{figure*}

Tables \ref{table:chall_bin} and \ref{table:chall_type} contain a performance comparison between our method and challenge methods for binary and type segmentation in MICCAI 2017 robotic instrument segmentation challenge. Our model prediction is submitted to the challenge portal by following the regulations to get the evaluation score. From the table \ref{table:chall_bin} and \ref{table:chall_type},  we can see that our model outperforms all other models by a significant margin for most of the data sequences. For type segmentation, team MIT obtains competitive results in sequences 5, 6, and 8, but the mean accuracy of our model achieves significant improvement. In Fig. \ref{fig:chal}, we compare the prediction of our model with the binary and instrument type segmentation of the participated models in the challenge \cite{allan20192017}. Our model's segmentation predictions consist of less false positive and false negative than other models of the challenge.

\section{Ablation Study}

\begin{table}[!h]
\centering
\caption{Our model performance and complexity while integrating different proposed methods and modules. Reg., Sal., and Scan. denote the MTL regularization, saliency, and scanpath. Tick and cross marks are represented the model performance with and without the corresponding modules respectively.}
\scalebox{0.92}{\begin{tabular}{c|c|c|c|c|c|c}
\hline
\multicolumn{4}{c|}{{Modules and Methods}} & {Seg.} & {{Sal.}} & \multirow{1}{.8cm}{{FPS}} \\ \cline{1-6}
\multirow{1}{1cm}{{ASTO}} & \multirow{1}{.8cm}{{SC-scSE}} & \multirow{1}{1.6cm}{{\begin{tabular}[c]{@{}c@{}}Conv-\\ LSTM++\end{tabular}}} & \multirow{1}{.7cm}{{\begin{tabular}[c]{@{}c@{}}Reg.\end{tabular}}} & {Type} & {Scan.} & \\ \cline{5-6} 
 &  &  &  & {Dice} & {Top-1} & {} \\ \hline
\cmark & \cmark & \cmark & \cmark & \textbf{0.773}  & 0.715  & 42 \\\hline
\xmark & \xmark & \cmark & \cmark & 0.385 & 0.687 & 42 \\ \hline
\cmark & \xmark & \cmark & \cmark & 0.671 & 0.692 & 56 \\\hline
\cmark & \cmark & \xmark & \cmark & 0.743 & 0.665 & 74 \\\hline
\cmark & \cmark & \cmark & \xmark & 0.768 & \textbf{0.750} & 42 \\ \hline
\cmark & \cmark & \xmark & \xmark & 0.768 & 0.657 & 74 \\ \hline
\cmark & \xmark & \xmark & \xmark & 0.657 & 0.665 & \textbf{100} \\ \hline

\end{tabular}}
\label{tab:module_ab}
\end{table}

\begin{figure}[!h]
    \centering
    \includegraphics[width=0.5\textwidth]{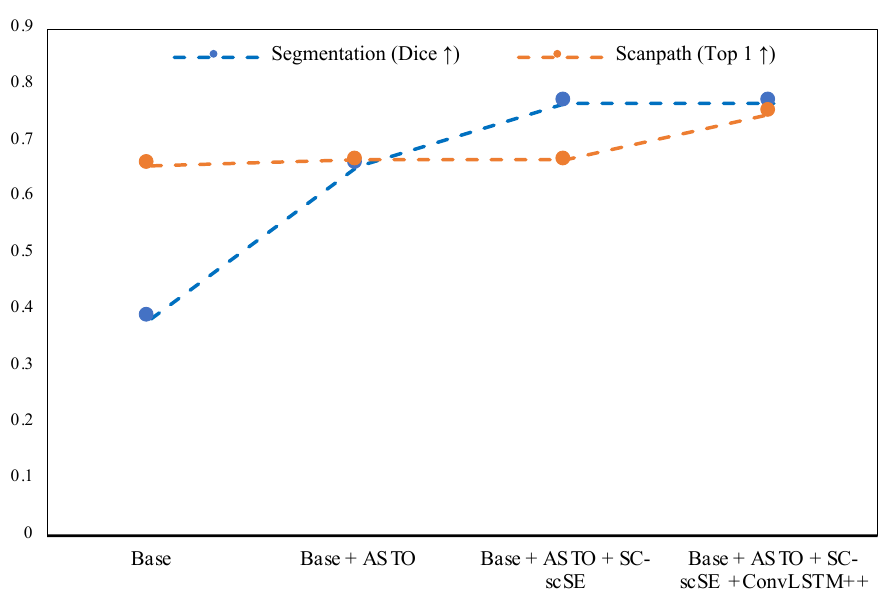}
    \caption{Performance improvement of our MTL model by integrating each proposed method and module. ASTO and SC-scSE show significant enhancement of the segmentation prediction where convLSTM++ improves the saliency detection in sequential video frames.}
    \label{fig:module_ablation}
\end{figure}

\begin{table}[!h]
\centering
\caption{Experimental comparison between scSE \cite{roy2018concurrent} and our attention proposed model. From results, we can infer that our attention module enhances segmentation prediction by a significant margin.}
\label{tab:ab_scse}
\begin{tabular}{l|c|c|c|c}
\hline
\multirow{1}{1cm}{{Modules}} & \multicolumn{2}{c|}{{Segmentation}} & \multicolumn{2}{c}{{Saliency}} \\ \cline{2-5} 
 & {Binary} & {Type} & {Saliency} & {Scanpath} \\ \cline{2-5} 
 & {Dice} & {Dice} & {BCE} & {Top-1} \\ \hline
{scSE} & 0.939 & 0.680 & 0.032 & 0.750 \\ \hline
{SC-scSE} & \textbf{0.955} & \textbf{0.768} & 0.032 & 0.750 \\ \hline
\end{tabular}
\end{table}

\begin{table}[!h]
\centering
\caption{Performance comparison of between ASTO and recent MTL optimization technique GradNorm \cite{chen2018gradnorm}. Our model requires spatio-temporal scheme of training which is not compatible with GradNorm mechanism.}
\label{tab:asto_gradnorm}
\begin{tabular}{l|c|c|c|c}
\hline
\multirow{1}{1cm}{{Modules}} & \multicolumn{2}{c|}{{Segmentation}} & \multicolumn{2}{c}{{Saliency}} \\ \cline{2-5} 
 & {Binary} & {Type} & {Saliency} & {Scanpath} \\ \cline{2-5} 
 & {Dice} & {Dice} & {BCE} & {Top-1} \\ \hline
{GradNorm} & 0.808 & 0.453 & 0.035 & 0.538 \\ \hline
{ASTO} & \textbf{0.961} & \textbf{0.773} & \textbf{0.032} & \textbf{0.715} \\ \hline
\end{tabular}
\end{table}

\begin{table}[!h]
\centering
\caption{ Quantitative analysis between ConvLSTM and proposed ConvLSTM++. From results, we can infer that ConvLSTM++ is capable of handling better long input sequences.}
\label{tab:ab_conv}
\scalebox{1}{\begin{tabular}{c|c|c|c|c|l}
\hline
\multirow {1}{1cm}{{Modules}} & \multicolumn{2}{c|}{{Seg.}} & \multicolumn{3}{c}{{Sal.}} \\ \cline{2-6} 
 & {Bin.} & {Type} & {Sal.} & \multicolumn{2}{c}{{Scanpath}} \\ \cline{2-6} 
 & {Dice} & {Dice} & {BCE} & {Top-1} & {Avg} \\ \hline
{ConvLSTM} & 0.955 & 0.768 & 0.033 & 0.705 & 0.672 \\ \hline
{Conv-LSTM++} & 0.955 & 0.768 & \textbf{0.032} & \textbf{0.750} & \textbf{0.721} \\ \hline
\end{tabular}}
\end{table}

\subsubsection{Module and Methods}
To validate our method and integration of the modules, we calculate the performance and complexity of the model, as shown in Table \ref{tab:module_ab}. The proposed optimization approach, ASTO, carries a high impact in our ST-MTL model for both segmentation and saliency estimation tasks. It resolves the common challenge of the MTL model to converge both tasks in the same epoch. We observe that the MTL model without ASTO converges the two tasks into different epochs such as segmentation at epoch-42 and saliency at epoch-67. Without ASTO, the convergence epoch of scanpath consists of poor segmentation accuracy in Table \ref{tab:module_ab}. ASTO achieves the convergence into the same epoch and retains the best accuracy for both tasks. \textit{Table \ref{tab:asto_gradnorm} represents the comparative performance of the ASTO over a recent MTL optimization technique, GradNorm \cite{chen2018gradnorm}. GradNorm is designed by focusing MTL spatial model whereas our approach requires spatio-temporal scheme of learning and obtains superior performance over GradNorm.} On the other hand, SC-scSE and ConvLSTM++ lift the segmentation and scanpath accuracy with a significant margin which can be inferred in Table \ref{tab:ab_scse} and Table \ref{tab:ab_conv}. The impact of each proposed module on our MTL model is demonstrated in Fig. \ref{fig:module_ablation}. However, ConvLSTM++ requires tremendous computational resources with larger learning parameters and remarkably reduces FPS's prediction speed (frame per second). Notably, MTL regularization slightly reduces the scanpath performance in-turn enhances the segmentation accuracy.

\subsubsection{Single-task and Multi-task}

\begin{table}[!h]
\centering
\caption{Performance comparison between single-task and multi-task model. Our multi-task model achieves higher performance than the models of individual tasks.}
\label{tab:single_ab}
\begin{tabular}{c|c|c|c|c|c|c}
\hline
\multicolumn{2}{c|}{{Task}} & \multicolumn{2}{c|}{{Segmentation}} & \multicolumn{3}{c}{{Saliency}} \\ \hline
\multirow{1}{.6cm}{{Seg.}} & \multirow{1}{.6cm}{{Sal.}} & {Bin.} & {Type} & {Sal.} & \multicolumn{2}{c}{{Scanpath}} \\ \cline{3-7} 
 &  & {Dice} & {Dice} & {BCE} & {Top-1} & {Avg.} \\ \hline
\cmark & \xmark & 0.939 & 0.680 & - & \textbf{-} & - \\ \hline
 \xmark & \cmark & - & - & 0.036 & 0.663 & 0.627 \\ \hline
\cmark & \cmark & \textbf{0.961} & \textbf{0.773} & \textbf{0.032} & \textbf{0.715} & \textbf{0.697} \\ \hline
\end{tabular}
\end{table}

Performance comparison between single-task and multi-task are shown in Table \ref{tab:single_ab}. We can observe that the MTL model improves both tasks compared to single-task models for segmentation and saliency. Our proposed ST-MTL optimization technique, ASTO, is playing vital rule behind the performance improvement. In ASTO, the shared encoder is learning the target instruments in the segmentation optimization. It facilitates the saliency detection of the same instruments in the phase of spatio-temporal optimization of saliency decoder. Moreover, MTL regularization refines the learning parameters for both tasks and attains the convergence point.

\subsubsection{Fusion loss and Regularization}

\begin{table}[!h]
\caption{Quantitative results for different fusion loss for task-oriented saliency loss and effect after multi-task regularization. Sink. denotes the Sinkhorn and bW is the short form of batch-Wasserstein loss \cite{islam2019learning}.}
\label{tab:fusion_ab}
\scalebox{1}{\begin{tabular}{c|c|c|c|c|c|c|c}
\hline
\multicolumn{3}{c|}{{Fusion Loss}} & \multicolumn{2}{c|}{{Segmentation}} & \multicolumn{3}{c}{{Saliency}} \\ \hline
\multirow{1}{.5cm}{{bW}} & \multirow{1}{.7cm}{{Sink.}} & \multirow{1}{.7cm}{{BCE}} & {Bin.} & {Type} & {Sal.} & \multicolumn{2}{c}{{Scanpath}} \\ \cline{4-8} 
&  &  & {Dice} & {Dice} & {BCE} & {Top-1} & {Avg.} \\ \hline
\cmark & \xmark & \cmark & 0.948 & 0.763 & \textbf{0.031} & \textbf{0.720} & \textbf{0.712} \\ \hline
\xmark & \cmark & \cmark & \textbf{0.961} & \textbf{0.773} & 0.032 & 0.715 & 0.697 \\ \hline
\end{tabular}}
\end{table}

In Table \ref{tab:fusion_ab}, we investigate the performance of the segmentation and scanpath with different fusion losses for the saliency detection and multi-task regularization. Though the fusion loss of the binary cross-entropy (\textit{BCE}) and batch-Wasserstein (\textit{bW}) \cite{islam2019learning} obtains the higher saliency and scanpath performance, it requires higher computation and reduces the segmentation accuracy after multi-task regularization. Therefore, we choose fusion loss of \textit{BCE} and Sinkhorn distance (\textit{Sd}) for our model. 

\subsubsection{Evaluation with SALICON}
 \begin{figure}[!h]
    \centering
    \includegraphics[width=0.5\textwidth]{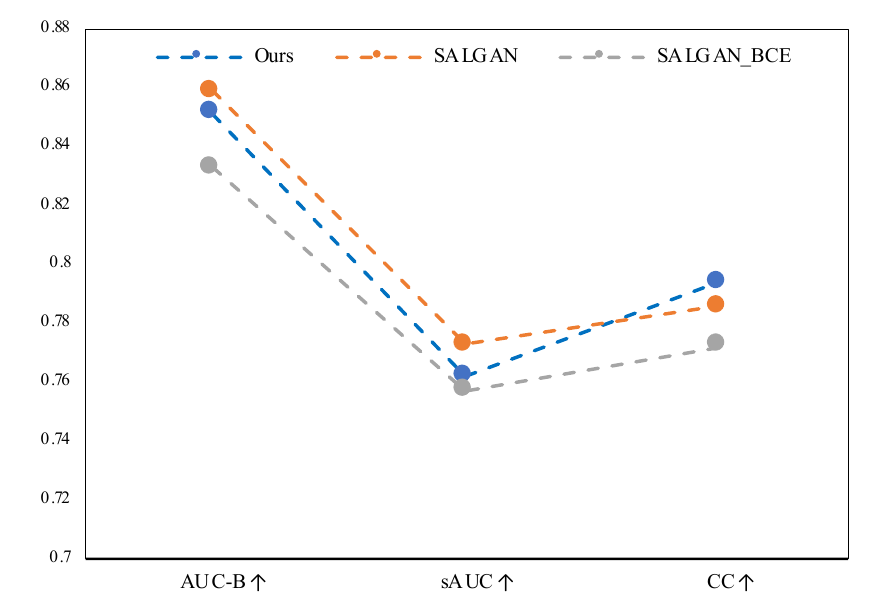}
    \caption{The performance comparison of our model over SalGAN \cite{pan2017salgan} and SalGAN\_BCE (with binary cross-entropy loss). We exploit evaluation metrics such as AUC-B (AUC Borji \cite{bylinskii2015saliency}), sAUC (AUC shuffle) and CC (Pearson's correlation coefficient) and the graph shows that our model achieves competitive performance in all the metrics. }
    \label{fig:salicon}
\end{figure}

The proposed saliency model is also validated with SALICON \cite{jiang2015salicon}, a state-of-the-art saliency dataset in computer vision. We replicate the same data setup as SalGAN \cite{pan2017salgan} to train our saliency model and evaluate the performance with the metrics of AUC-B (AUC Borji \cite{bylinskii2015saliency}), sAUC (AUC shuffle) and CC (Pearson's correlation coefficient). As SALICON does not consist of temporal data (static images rather than video sequences as our robotic instrument dataset), we ignore the convLSTM++ module conduct all the experiments with it. Fig \ref{fig:salicon} demonstrates the prediction comparison of our model over SalGAN \cite{pan2017salgan} and SalGAN\_BCE (with binary cross-entropy loss). Though our model design in consideration of temporal dependency, the performance is still competitive comparing to other saliency models.

\section{Discussion}

From the quantitative and qualitative results, we can summarize that the proposed ST-MTL model outperforms all the SOTA models of segmentation and saliency and participant models in the robotic instrument segmentation challenge \cite{allan20192017}. In the Table  \ref{tab:module_ab} and \ref{tab:single_ab}, it is clear that ASTO is capable to resolve the multi-task convergence problem of the MTL model optimization. In our case, ASTO works perfectly as the target objects are the same for both segmentation and saliency tasks. The trained weight of shared-encoder with segmentation assists saliency decoder to optimize independently. Since ASTO generates the task-specific gradients with corresponding losses, it can also be applied for an exclusive spatial or temporal MTL model. Besides, SC-scSE and ConvLSTM++ improve the model performance significantly by boosting feature excitation and long-range spatio-temporal correlation, respectively. However, ConvLSTM++ is computationally expensive, with the large learning parameters of \textit{28.31 Millions}. We observe that adding the ConvLSTM++ module with the segmentation decoder does not improve the accuracy rather degrades the prediction frame rate. On the other hand, we apply computation efficiency, and GPU enable fusion loss of BCE and Sinkhorn distance (Sd) for saliency task, despite batch-Wasserstein (bW) \cite{islam2019learning} obtains better performance in the scanpath estimation (see the Table \ref{tab:fusion_ab}). \textit{Though the performances of the ST-MTL are compared with the single task models in Tables \ref{table:chall_bin} and \ref{table:chall_type} with the segmentation task, the MTL model takes advantage of additional features with the shared encoder from the saliency task. }

\section{Conclusion}
In this paper, we propose an end-to-end trainable Spatio-Temporal Multi-Task Learning (ST-MTL) model for real-time surgical instrument segmentation and task-specific attention prediction in the robot-assisted surgical scenario. The model can guide the endoscope movements similar to the surgeon's visual perception while tracking instruments in robotic surgery. We introduce a novel Asynchronous Spatio-Temporal Optimization (ASTO) to optimize the ST-MTL model for shared-encoder and task-aware decoders. ASTO rectifies the multi-task convergence issue by optimizing the decoders separately with shared-encoder. We also design skip competitive scSE (SC-scSE) module with skip connection to differentiate weak and strong features significantly. To capture long-range spatio-temporal correlation, we construct ConvLSTM++ by concatenating the encoder feature of the present frame with the previous frame, which boosts task-oriented saliency performance. Our proposed ST-MTL model outperforms the overall winner of MICCAI 2017 Robotic Instrument Segmentation challenge by a significant margin in both binary and type segmentation (see the Table \ref{table:chall_bin} and \ref{table:chall_type}).

The image-guided robotic surgical system should understand the complete surgical scene with the capability of tracking, detection, visual perception, and multi-tasking, similar to a human surgeon. Segmentation of defected tissues with instruments, workflow estimation, pose estimation, and interaction classification can be the future direction of the MTL model in robot-assisted surgery.

\section*{Acknowledgments}
This work was supported by the Singapore Academic Research Fund under Grant {R-397-000-297-114}, National Key Research and Development Program, The Ministry of Science and Technology (MOST) of China (No. 2018YFB1307703) and NMRC Bedside \& Bench under grant R-397-000-245-511.

\bibliographystyle{IEEEtran}
\bibliography{refs}

\end{document}